# Cognitively-Inspired Episodic Memory Architectures for Accurate and Efficient Character AI

Rafael Arias Gonzalez    and    Steve DiPaola

Simon Fraser University
sdipaola@sfu.ca

Large language models show promise for embodying historical characters in dialogue systems, but existing approaches face a critical trade-off: simple retrieval-augmented generation produces shallow responses, while multi-stage reflection achieves depth at prohibitive latency. We present an architecture that resolves this tension through offline data augmentation and efficient parallel retrieval from structured episodic memory. Our system transforms biographical data into 1,774 enriched first-person memories with affective-semantic metadata, then employs two-stage retrieval achieving 0.52s prompt generation. Evaluation using LLM-as-judge and RAGAs metrics shows our approach achieves parity with traditional RAG on GPT-4 while significantly outperforming it on smaller models (GPT-3.5, GPT-3), suggesting particular value for resource-constrained deployments. Beyond dialogue, the structured memory enables novel visualization tools: spatiotemporal heatmaps, emotional trajectory analysis, and interactive path tracking, positioning the system as both a dialogue interface and research tool for biographical analysis. We use Van Gogh as a test case, but the architecture is generalizable to any historical figure with substantial textual records, offering a practical framework for educational, museum, and research applications requiring both accuracy and efficiency.

## CCS Concepts

• Computing methodologies → Knowledge representation and reasoning;

• Computing methodologies → Natural language generation;

• Human-centered computing → Human computer interaction (HCI);

**Additional Keywords and Phrases:** retrieval-augmented generation, episodic memory, cognitive architectures, conversational AI, character embodiment, educational AI, museum applications

## 1  INTRODUCTION

Creating believable AI character agents has evolved from rule-based systems to contemporary LLM implementations, yet fundamental challenges persist. Historical figures present a particularly compelling use case. Their lives are well-documented through biographies, letters, and archival materials, yet traditional approaches to character embodiment struggle to transform this rich textual heritage into coherent, interactive experiences.

Recent advances in LLMs have renewed interest in character-based dialogue systems. However, when deployed without grounding mechanisms, these systems exhibit a persistent tendency toward hallucination, generating plausible but factually incorrect information with high confidence. For historical figures like Vincent van Gogh, where accuracy matters for educational and cultural preservation applications, such fabrications undermine system trustworthiness and limit practical deployment.

Current approaches to character embodiment face a fundamental trade-off between accuracy and responsiveness. Simple retrieval-augmented generation (RAG) systems can ground responses in factual data but often produce shallow, disconnected answers that fail to capture the experiential richness of a character's lived history. Multi-stage architectures incorporating self-reflection, iterative refinement, and chained retrievals achieve greater depth and coherence but at severe computational cost. Systems requiring multiple sequential LLM calls can take tens of seconds to construct a single response, far too slow for natural conversation and prohibitively expensive for real-time educational applications. Our work addresses a gap in conversational AI: existing systems optimize either for accuracy (through expensive multi-stage processing) or efficiency (through simple retrieval), but educational and museum deployments require both. By conducting reflection and enrichment offline, we achieve the contextual depth of multi-stage systems at single-retrieval latency. This design philosophy—separating concerns temporally rather than eliminating them—offers a generalizable approach to resource-constrained AI deployment.

This latency problem becomes particularly acute when considering deployment constraints. While large-scale models like GPT-4 and Claude v4 with large context windows offer impressive capabilities, educational institutions, museums, and research environments often require systems that can run locally or on modest cloud infrastructure. Smaller models that enable such deployment show even greater performance degradation with traditional RAG approaches, making the efficiency-accuracy gap a critical barrier to practical application.

We present an architecture that resolves the efficiency-accuracy dilemma through a novel combination of offline data augmentation and real-time parallel retrieval. Our key insight is that the computationally expensive work of self-reflection, context enrichment, and perspective transformation can be performed once during dataset construction rather than repeatedly during each interaction. By pre-processing biographical and epistolary data into structured episodic memories enriched with affective-semantic metadata (temporal markers, emotional valence/arousal, character relationships, geographic locations), we create a substrate that enables both rapid retrieval and rich contextualization.

Our system implements a three-tier memory architecture inspired by cognitive models:

1. **Long-term episodic memory**:1,774 first-person experiential memories derived from biographic text letters
2. **Intermediate conversational memory**: dynamically maintained records of recent dialogue
3. **Short-term working memory** (the LLM's context window).

A two-stage retrieval mechanism accesses these memory tiers in parallel. Initial similarity search uses concise memory descriptions optimized for matching user queries, then retrieves full first-person narratives and associated affective-semantic metadata for prompt construction. The result is a system achieving an average prompt generation time of 0.52 seconds while maintaining response quality that outperforms traditional RAG implementations, particularly for smaller, more deployable models. Our evaluation using LLM-as-a-judge and RAGAs (RAG assessment) frameworks demonstrates consistent advantages in faithfulness and contextual relevance across multiple model scales.

While conversational interaction represents the most visible application of our architecture, the structured nature of our episodic memory dataset enables a complementary use case: interactive visualization and exploration of historical figures' cognitive and experiential landscapes. Our affective-semantic enrichment (valence, arousal, geographic coordinates, character co-occurrence, autobiographical significance) transforms biographical data into a multidimensional space that can be analyzed, visualized, and navigated. We demonstrate applications including spatiotemporal heatmaps, emotional trajectory analysis, social network visualization, and user path tracking through memory space during conversations. This dual-use positioning distinguishes our work from pure chatbot systems, serving both real-time dialogue generation and contemplative historical investigation. We demonstrate our approach using Vincent van Gogh, whose extensive documentation provides ideal source material but represents a generalizable case.

We contribute three advances to conversational AI for character embodiment:

1. An efficient episodic memory architecture achieving sub-second prompt generation (0.52s average) through offline augmentation and parallel retrieval, enabling real-time dialogue on resource-constrained hardware.

2. A multi-stage augmentation methodology transforming unstructured biographical materials into structured first-person memories with affective-semantic metadata (valence, arousal, geocoding, temporal markers).

3. Empirical evidence that structured episodic memory democratizes character embodiment: while achieving parity with traditional RAG on large models (GPT-4), our approach significantly outperforms it on smaller models (GPT-3.5, GPT-3), with implications for sustainable and accessible AI deployment.

## 2 RELATED WORK

We build on four research threads: character embodiment, memory architectures, RAG systems, and LLM-driven data augmentation.

**Character Embodiment and Role-Playing Systems:** Recent work on character embodiment has explored various approaches to creating believable AI agents. Park et al.'s (2023) Generative Agents demonstrated sophisticated behavioral simulation through memory streams and reflection mechanisms, though at substantial computational cost due to continuous LLM invocations. Character-LLM (Shao et al., 2023) achieves strong performance through fine-tuning character-specific models on first-person narratives, but requires separate training for each character and retraining for knowledge updates. They acknowledge that relying primarily on Wikipedia limits biographical depth compared to comprehensive sources. RoleLLM (Wang et al., 2023) emphasizes rich character descriptions in prompts and notes that explicitly instructing models to reference context directly yields better responses. Li et al.'s (2023) ChatHaruhi similarly highlights the importance of personality cues and conversational history retrieval. Unlike fine-tuning approaches requiring per-character model training, our retrieval-based architecture enables rapid character instantiation through dataset construction alone. Compared to continuous reflection systems, we achieve equivalent depth through offline processing, eliminating runtime computational overhead.

**Memory Systems and Cognitive Architectures:** Memory systems in AI agents draw inspiration from cognitive science models distinguishing short-term, working, and long-term memory (Cowan, 2008), with Tulving's (1972) distinction between semantic and episodic memory providing theoretical grounding. Recent implementations include Kim et al.'s (2023) memory-augmented LLM personalization and Zhang et al.'s (2023) user modeling systems, which focus on learning individual preferences rather than instantiating pre-defined historical characters. Sant'Anna et al.'s (2023) work on autonoetic consciousness (subjective re-experiencing of past events) has received limited attention in AI systems despite its relevance to episodic memory. We explicitly model three memory tiers inspired by cognitive architectures while grounding long-term memory in structured episodic representations enriched with affective-semantic metadata.

**RAG Architectures:** Retrieval-augmented generation (Lewis et al., 2020) grounds LLM outputs in external knowledge. Gao et al.'s (2024) survey categorizes RAG systems from naive single-retrieval to sophisticated modular architectures, noting that while modular designs offer adaptability, they incur substantial computational cost through multiple LLM invocations. Shuster et al. (2021) found that RAG substantially improves factual accuracy in conversational AI, particularly relevant for character embodiment requiring biographical fidelity. Recent advances in complex RAG patterns including self-reflection, tool-interactive critiquing, and iterative self-refinement (Ji et al., 2023; Gou et al., 2024; Madaan et al., 2023) enhance output quality but introduce latency incompatible with real-time interaction. We achieve RAG-level grounding while approaching naive RAG speed through offline preprocessing. By conducting "self-reflection" during dataset construction rather than runtime, we obtain the contextual depth of multi-stage systems at single-retrieval latency.

**Efficiency and Model Scaling Considerations:** The tension between model capability and deployment practicality has received growing attention. Strubell et al. (2019), Desislavov et al. (2021), and Thompson et al.(2021) document substantial energy costs and diminishing returns in large model development. The emergence of smaller models like Gemma 2B/7B, Llama 3.1 8B, and Mistral 7B reflects industry recognition of deployment constraints (Gemma Team et al., 2024; Meta, 2024; Mistral AI, 2024). Our design specifically targets deployment on smaller models through strict token budgets (2000 tokens total) and optimized retrieval efficiency, anticipating edge deployment scenarios where computational resources, energy consumption, and inference costs require careful management.

**LLMs as Data Augmentation Tools**: These have gained traction across NLP tasks, from conversational reformulations (Chen et al., 2024) to few-shot entity recognition (Ye et al., 2024) and multilingual augmentation (Whitehouse et al., 2023). In HCI contexts, LLM-generated synthetic data for user studies (Hämäläinen et al., 2023; Miotto et al., 2022) focus primarily on generating new participants rather than transforming existing biographical materials. We contribute a multi-stage augmentation pipeline specifically designed for biographical data transformation, systematically converting third-person narratives to first-person perspectives (autonoetic) while extracting affective-semantic metadata.

**Evaluation Methodologies:** Evaluating character-based dialogue systems presents challenges, as traditional metrics like BLEU and ROUGE poorly capture emotional authenticity and perspective consistency (Gao et al., 2024). Zheng et al.'s (2023) LLM-as-a-judge framework achieves 80%+ agreement with human evaluators while identifying key biases (position, self-enhancement, verbosity). Es et al.'s (2023) RAGAs framework provides reference-free metrics for RAG systems: Context Relevance, Answer Relevance, and Faithfulness. We employ both methodologies complementarily, enabling comparative evaluation against baselines while assessing retrieval and grounding mechanisms without requiring extensive human annotation.

Across these threads, a gap remains between deep, reflective architectures that are too slow for interaction and fast RAG pipelines that lack experiential richness. By performing reflection offline and encoding results as structured memories, our approach delivers depth, grounding, and responsiveness in a single design.

# 3  SYSTEM ARCHITECTURE

Our architecture addresses the efficiency-accuracy dilemma through a principled separation of concerns: computationally expensive operations occur once during offline dataset construction, while runtime operations prioritize speed through parallel retrieval and structured access patterns. This design enables sub-second prompt generation while maintaining the contextual richness typically associated with multi-stage reflection systems. Figure 1 provides an overview of the complete system, showing the relationship between offline augmentation, runtime retrieval, and dialogue generation.

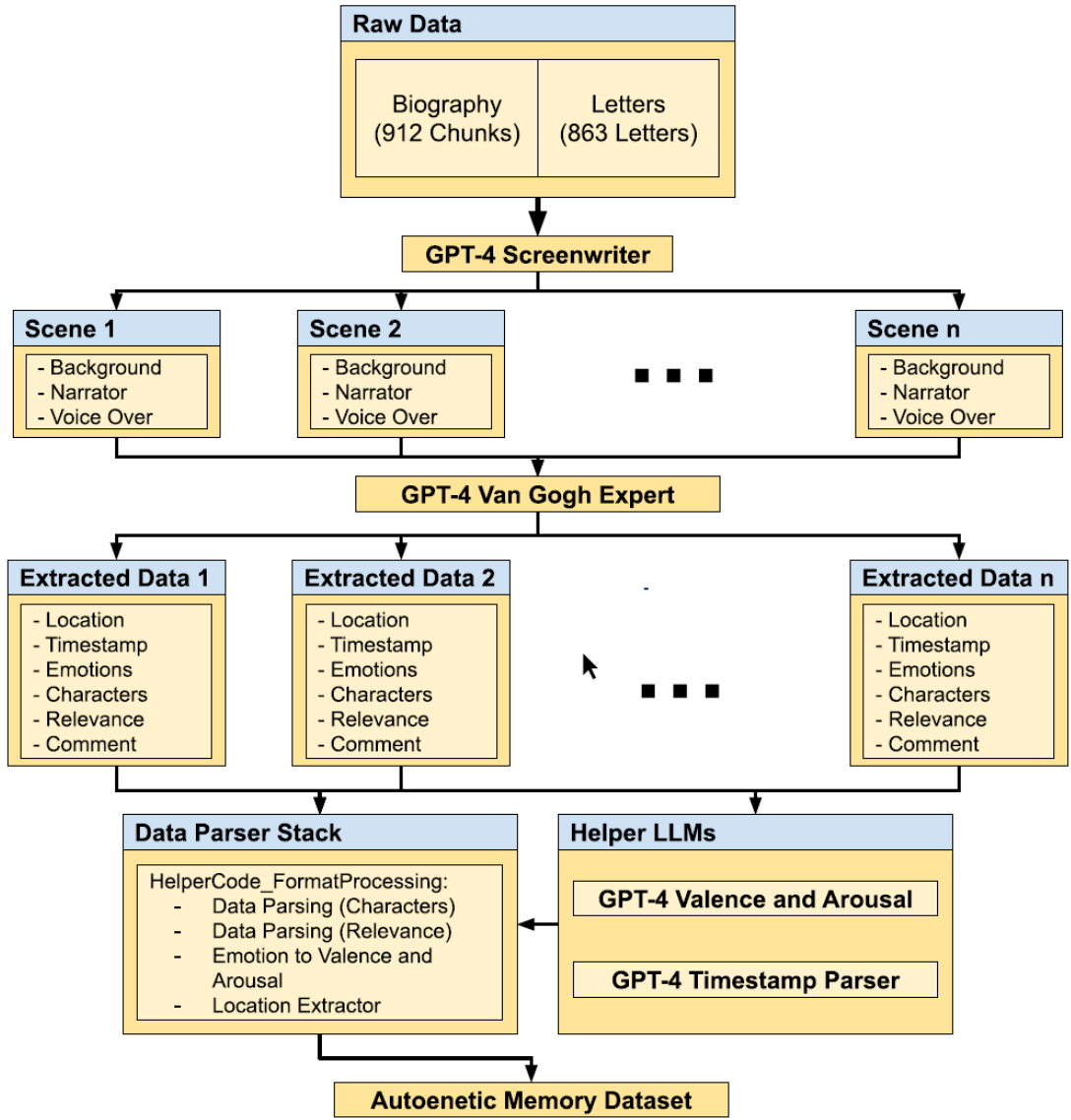

Figure 1. Autonoetic Dataset Pipeline.

## 3.1 Design Philosophy and Generalizability

The architecture comprises four main stages:

1. **Offline Data Augmentation**: Multi-stage LLM processing transforms source materials into structured episodic memories enriched with affective-semantic metadata
2. **Memory Indexing**: Processed memories are embedded and organized into retrievable indexes
3. **Runtime Retrieval**: User queries trigger parallel search across memory tiers with two-stage access

4. **Prompt Construction and Generation**: Retrieved memories and metadata populate LLM context for response generation

This separation between offline and online processing proves critical for efficiency. Operations requiring iterative LLM calls, information extraction, and data enrichment occur once during dataset construction. Runtime operations reduce to embedding similarity search and prompt assembly, both computationally inexpensive operations achievable in sub-second timeframes. The system requires only two inputs: (1) substantial textual materials about the character (biography, letters, interviews, memoirs), and (2) character-defining attributes for static context (personality traits, communication style, historical period).

## 3.2 Three-Tier Memory Architecture

Inspired by cognitive models distinguishing memory types by temporal scope and accessibility (Cowan, 2008), our system implements three memory tiers that parallel human memory organization while remaining computationally tractable.

### 3.2.1 Long-Term Episodic Memory

The foundation of our system is a structured dataset of first-person experiential memories derived from biographical source materials. Each memory entry contains:

- **Narrative content**: First-person account of an experience, including setting, actions, thoughts, and emotions
- **Temporal markers**: Dates situating the memory in the character's lifespan
- **Affective metadata**: Emotional valence (pleasant/unpleasant dimension) and arousal (activation level)
- **Social context**: Characters present or mentioned in the memory
- **Geographic data**: Locations as geocoded coordinates
- **Relevance scores**: Autobiographical significance ratings

For our Van Gogh implementation, this dataset comprises 1,774 memories spanning 1853-1890, with particularly dense coverage during his artistically productive periods (1880-1890). The dataset structure (Table 1 in Section 4) generalizes to any historical figure; the fields remain consistent while content varies by individual. Long-term memory serves dual purposes: it provides retrieval content for dialogue generation and enables analytical visualization of the character's experiential landscape.

### 3.2.2 Intermediate and Short Term Conversational Memory

While long-term episodic memory remains static (representing the character's fixed biographical history), conversational exchanges generate new experiential data requiring retention for thematic continuity. We implement intermediate memory through dynamic accumulation of query-response pairs embedded into a secondary retrieval index.

After each dialogue turn, the system concatenates the user query and LLM response into a single text unit, generates an embedding representation, and stores it in a second retrieval index (RAG2) alongside the original long-term memory embeddings. Subsequent queries trigger similarity search across both long-term biographical memories and recent conversational exchanges. If conversational memory yields high-similarity results, the system performs a secondary retrieval using the conversation memory as a query against long-term memory, establishing explicit connections between current discussion topics and relevant biographical experiences.

This mechanism enables thematic continuity without expanding context window requirements. Intermediate memory exists only within individual conversation sessions. Resetting the chatbot clears conversational memory while preserving long-term biographical memory, analogous to starting a new interview with the same historical figure. The LLM's context window serves as short-term working memory, holding immediately active information for response generation. Our implementation maintains strict token budgets to ensure compatibility with smaller models and future edge deployment with a 2000 token total:.

- Retrieved long-term memories: 600 tokens;
- Response generation: 500 tokens
- Static context: 300 tokens
- Affective-semantic metadata: 100 tokens;
- Recent conversation history: 500 tokens

This conservative allocation reflects our target deployment context. While contemporary models support large token windows, smaller models suitable for local deployment (Gemma 7B, Llama 3 8B, Mistral 7B) typically offer 2k-8k contexts. Our 2k budget ensures forward compatibility with emerging edge-deployable models while enforcing discipline in information selection.

### 3.3 Two-Stage Retrieval Mechanism

A key architectural innovation is our two-stage retrieval process, which separates the representation used for similarity matching from the content provided to the LLM for generation. This design resolves a fundamental tension: concise descriptions enable better query matching, while rich first-person narratives enable expressive response generation.

**Stage One:** Similarity-Based Retrieval:  When a user poses a query, the system embeds the query using the same embedding model employed for memory indexing (OpenAI's text-embedding-ada-002), computes cosine similarity between the query embedding and all memory embeddings, ranks memories by similarity score, and retrieves top-k candidates (k=3 in our implementation).

The embeddings represent not the full first-person narratives but augmented context descriptions, condensed summaries capturing the core event, temporal reference, key characters, and primary location. For example: "Argument with brother Theo about financial support and artistic direction, December 1881, The Hague. Present: Theo van Gogh (mentioned), parents (mentioned)." This concise representation aligns better with typical user queries, which often reference topics, time periods, or relationships rather than matching the narrative style of first-person memories.

**Stage Two:** Content Retrieval and Enrichment: Having identified relevant memories through similarity search, the system retrieves their complete content for prompt construction: first-person narrative, temporal markers, affective metadata (valence/arousal values), social context (characters present or mentioned), geographic data, and relevance score. This two-stage approach (searching on concise descriptions but generating from rich narratives) optimizes both retrieval precision and expressive capacity.

### 3.4 Parallel RAG Architecture

Our system employs two Retrieval-Augmented Generation pipelines operating in parallel, both querying the same underlying episodic memory dataset but serving different retrieval purposes. RAG1 (Long-term Memory Access) directly retrieves episodic memories based on query similarity. RAG2 (Conversational Memory + Associated Episodics) maintains embeddings of query-response pairs from the current conversation. When conversational memory retrieval succeeds, RAG2 performs a secondary retrieval using the conversation memory as a query against the long-term episodic dataset.

This parallelization enables efficient thematic continuity without sequential dependency. If RAG2 finds relevant conversational memory, it retrieves one associated long-term memory, allowing the final prompt to include: 1 conversational memory (showing topical continuity), 1 long-term memory related to that conversation, and 1-2 long-term memories directly matching the current query. If no conversational memory meets the similarity threshold, the system defaults to retrieving 3 long-term memories directly relevant to the current query. Figure 2 shows the complete retrieval process.

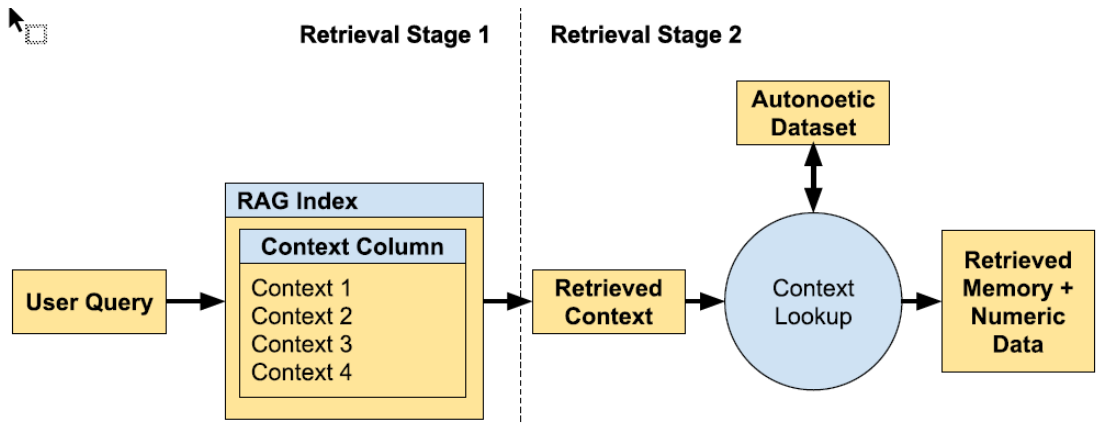

Figure 2. Our 2-Stage Retrieval Process.

## 3.5 Prompt Construction Strategy

Retrieved memories and metadata populate the LLM's context window following a structured template. The prompt includes: static character context (300 tokens) with personality traits and role-playing instructions; retrieved episodic memories (600 tokens) with first-person narratives, dates, and locations; affective-semantic metadata (100 tokens) including emotional state, characters present, and relevance scores; and recent conversation history (500 tokens). This structure provides the LLM with multiple context layers: stable character definition, specific relevant experiences, emotional and social context, and conversational continuity.

## 3.6 Efficiency Analysis

The architecture's efficiency derives from three design decisions. First, offline complexity, online simplicity: expensive operations (perspective transformation, information extraction, emotional annotation) occur once during dataset construction, while runtime operations reduce to similarity search and prompt assembly. Second, parallelization without sequential dependency: both RAG pipelines query simultaneously rather than sequentially, maintaining constant time complexity. Third, pre-computed embeddings: by caching embeddings for all long-term memories and reusing them across sessions, we eliminate repeated API calls.

The result is an average prompt generation time of 0.52 seconds measured across hundreds of test interactions. At typical GPT-4 generation speeds (~5-7 tokens/second), the model would generate only 3-4 words in the time our system constructs a complete, contextually rich 1500-token prompt. Multi-stage reflection systems requiring 3-5 sequential LLM calls would require 15-30+ seconds for equivalent complexity, a latency incompatible with conversational interaction expectations.

Our modular implementation uses OpenAI APIs with support for local models via Hugging Face transformers, enabling seamless transition from cloud-based prototyping to local deployment.

# 4   MULTI-STAGE DATASET CONSTRUCTION

This section describes our methodology for transforming unstructured biographical materials into structured episodic memory datasets enriched with affective-semantic metadata. While we demonstrate the approach using Vincent van Gogh as our exemplar, the pipeline generalizes to any historical figure with substantial textual documentation.

## 4.1 Source Material Requirements and Selection

Our approach requires two types of source materials: biographical narratives (third-person accounts of the figure's life) and first-person documents (letters, diaries, memoirs, interviews). The combination proves essential since biographies provide chronological structure and contextual framing, while first-person materials reveal subjective perspective and authentic voice.

**Biography**: We used Naifeh and Smith's *Van Gogh: The Life* (2011), a comprehensive 976-page biography created in collaboration with the Van Gogh Museum and incorporating previously unpublished archival materials. This work spans Van Gogh's complete lifespan (1853-1890) with particular depth during his artistic period (1880-1890). The biography was segmented into approximately 912 chunks of ~3000 characters each (roughly 3 paragraphs), balancing the need to capture contextual richness while representing distinct episodic experiences.

**Letters**: Van Gogh's correspondence comprises 863 letters written both by and to him. These documents provide direct access to his thoughts, emotional states, and subjective interpretations of life events. However, they present significant processing challenges: fragmented structure, assumed contextual knowledge, and lack of explicit temporal markers and location specificity that would enable effective retrieval.

For generalization to other historical figures, our methodology requires: comprehensive third-person biographical coverage spanning significant life periods; first-person materials revealing subjective perspective (optional but highly beneficial); and texts in machine-readable format. Figures meeting these criteria include scientists, political leaders, authors, and artists with documented textual legacies. The pipeline relies on general-purpose LLM capabilities applicable across biographical contexts.

## 4.2 Multi-Stage Augmentation Pipeline

Figure 3 illustrates our complete augmentation pipeline, comprising four sequential LLM-driven stages followed by automated enrichment and quality assurance.

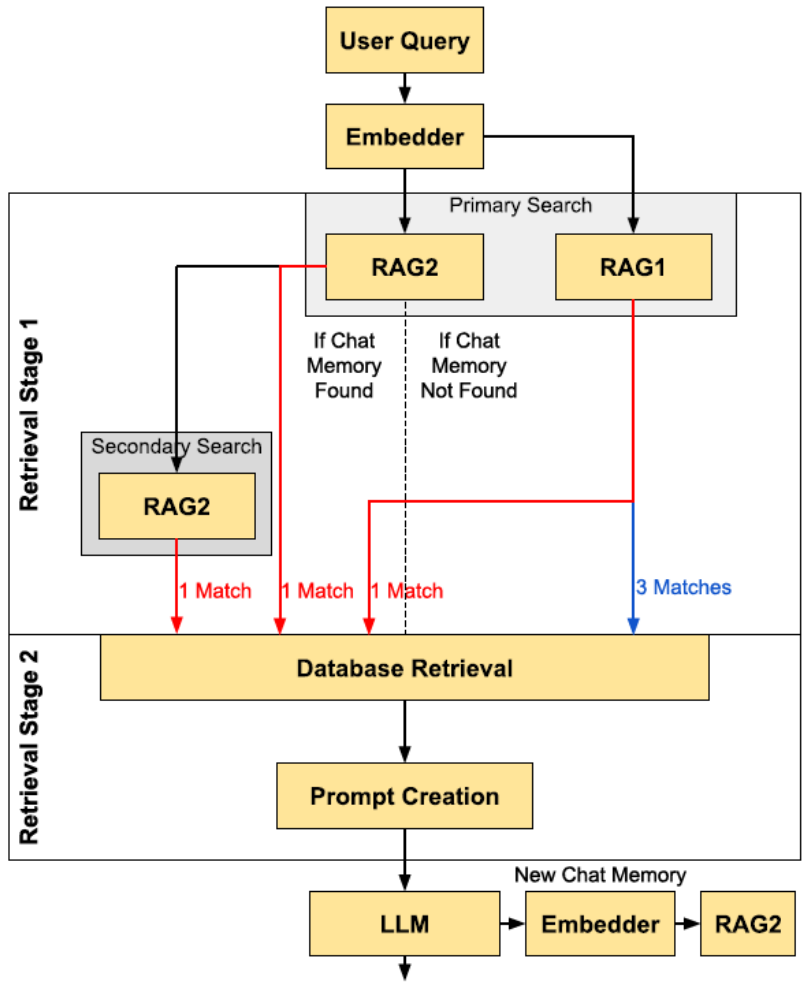

Figure 3. Multi-Stage Primary and Secondary Search

### 4.2.1 Stage 1: Perspective Transformation via Screenplay Generation

**Objective**: Convert third-person biographical narratives and fragmented first-person letters into coherent first-person experiential accounts while preserving factual content.

**Method**: We employ an LLM in a specialized "screenwriter" role, instructed to transform source text into movie script format featuring: NARRATOR (V.O.) for brief contextual framing, BACKGROUND for scene description, and [CHARACTER NAME] (V.O.) for first-person voiceover narration capturing subjective experience.

**Example**: *Original*: "In summer 1888, Vincent experienced isolation in Arles. Theo urged him to moderate his painting pace." *Transformed*: "VINCENT VAN GOGH (V.O.): My brother Theo pleads for slower strokes. But how can I deny the fervor that fuels my existence?"

### 4.2.2 Stage 2: Information Extraction and Contextualization

**Objective**: Systematically extract structured information from transformed narratives to enable both retrieval optimization and analytical visualization.

**Method**: A second LLM adopts a "Van Gogh biographer expert" role, analyzing each screenplay scene to extract: Location (standardized address format), Timestamp (DD/MM/YYYY format), Emotions (single-word labels), Characters (present or mentioned), Context (one-sentence explanation), Relevance (1-10 scale), and Comments (brief expert analysis).

### 4.2.3 Stage 3: Affective-Semantic Metadata Enrichment

We perform quantitative enrichment to enable analytical operations. We adopt the dimensional model of emotion (Heilman, 1997), representing affective states along two continuous axes: Valence (pleasantness, -1 to +1) and Arousal (activation level, -1 to +1). For each memory, we aggregate all unique emotion labels, assign valence/arousal values to each emotion, then average these values. Locations are geocoded to latitude/longitude coordinates using the OpenCage Geocoding API, enabling spatiotemporal visualization. Dates in various formats are normalized to Pandas Timestamp objects, enabling chronological sorting and time-series analysis.

### 4.2.4 Stage 4: Augmented Context Generation

For retrieval optimization, we construct **augmented context descriptions**, concise summaries combining the biographer's context statement with key metadata. Format: "[Context sentence]. [Date]. [Location]. Characters: [list]." These descriptions serve as embedded representations for similarity search while full first-person narratives populate prompts for generation.

### 4.3 Quality Assurance and Error Correction

LLM-driven augmentation introduces potential inconsistencies requiring systematic correction. Manual review identified 14 instances (<1% of total) where the LLM assigned dates outside Van Gogh's lifespan, requiring correction. We developed Python scripts with regex patterns to parse and normalize all extracted fields automatically. Some location descriptions proved too vague for automated geocoding, affecting ~5% of memories.

### 4.4 Final Dataset Structure and Statistics

Table 1 presents the complete dataset schema. Each of the 1,774 memories contains:

| Field | Description | Data Type |
|---|---|---|
| UID | Unique identifier | String |
| Background | Scene setting description | Text |
| Narrator | Contextual framing statement | Text |
| Van Gogh Voiceover | First-person experiential narrative | Text |
| Context | One-sentence biographical context | Text |
| Comment | Expert biographer analysis | Text |
| Characters | Individuals present or mentioned | List[String] |
| Valence | Emotional pleasantness dimension | Float [-1, 1] |
| Arousal | Emotional activation dimension | Float [-1, 1] |
| Timestamp | Chronological marker | DateTime |
| Latitude | Geographic coordinate | Float |
| Longitude | Geographic coordinate | Float |
| Relevance | Autobiographical significance | Integer [1-10] |
| Augmented_Context | Concise summary for retrieval | Text |

**Table 1**. Structured episodic memory dataset schema.

**Dataset Statistics**: Total memories: 1,774 (913 from biography, 861 from letters); Temporal coverage: 1853-1890 (37 years); Mean valence: -0.09 (SD=0.45); Mean arousal: 0.40 (SD=0.25); Characters documented: 247 unique individuals; Geographic locations: 52 distinct places across 8 countries.

### 4.5 Generalization Strategy

To adapt this pipeline for different historical figures: replace source materials, modify character name in screenplay prompts, adjust expert role to "[Figure Name] biographer," and maintain the same schema. The pipeline logic, processing stages, and technical infrastructure require no modification. The approach requires source materials in substantial quantities (minimum ~500 pages recommended), adequate LLM knowledge about the figure, and first-person materials preserving authentic voice where available.

## 5 EVALUATION

We evaluate our architecture across two primary dimensions: efficiency (prompt generation speed enabling real-time interaction) and quality (response accuracy, faithfulness, and contextual relevance). Our evaluation strategy employs complementary methodologies (computational timing measurements, LLM-as-a-judge comparative analysis, and reference-free RAGAs metrics) to provide comprehensive assessment without requiring extensive human annotation.

### 5.1 Research Questions

Our evaluation addresses three core questions:

**RQ1 (Efficiency)**: Does offline augmentation with parallel retrieval achieve prompt generation speeds compatible with real-time conversational interaction?

**RQ2 (Quality - Comparative)**: How does structured episodic memory retrieval compare to traditional RAG implementations in terms of response quality across different model scales?

**RQ3 (Quality - Absolute)**: Do our retrieval mechanisms provide faithful, contextually relevant information that grounds LLM responses in biographical source materials?

These questions reflect our dual objectives: demonstrating practical viability for deployment (RQ1) and validating that efficiency gains don't compromise response quality (RQ2-3).

### 5.2 Experimental Setup

### 5.2.1 Models and Baselines

We evaluated three LLM variants to assess performance across model scales:

- **GPT-4** (8k context window): Representing strong baseline performance at time of evaluation
- **GPT-3.5-turbo** (4k context window): Mid-tier model representative of current cloud deployment
- **GPT-3 Davinci-002** (4k context window): Legacy model approximating zero-shot performance of smaller deployable models (Gemma 7B, Llama 3 8B)

**Baseline System (Traditional RAG)**: A standard retrieval-augmented generation implementation using the same static character context but retrieving directly from raw biographical chunks and letter text. This baseline lacks first-person transformation, affective-semantic metadata, and the two-stage retrieval mechanism.

**Our System Variants**: We tested three configurations in ablation studies:

1. **Static Only**: Character context without any retrieval (baseline establishing zero-augmentation performance)
2. **Static + Episodic**: Character context plus episodic memory retrieval, but without affective-semantic metadata
3. **Full System**: Complete architecture including static context, episodic memory retrieval, and affective-semantic metadata enrichment

### 5.2.2 Test Sets

**Question Set 1 (LLM-as-a-Judge)**: We employed GPT-4 to generate 30 objective questions about Van Gogh's life, avoiding potential authorship bias. We supplemented these with 3 manually crafted queries targeting specific aspects: the ear incident, precise date queries, and a deliberately uneventful date (testing hallucination resistance). Total: 33 questions.

**Question Set 2 (RAGAs)**: A separate set of 60 diverse questions developed within our research group, covering historical events, artistic development, emotional experiences, and relationship dynamics. This set emphasizes biographical breadth rather than specific factual tests. Total: 60 questions.

### 5.2.3 Evaluation Frameworks

**LLM-as-a-Judge (Zheng et al., 2023)**: We adapted this framework for pairwise comparison, using GPT-4 as an impartial evaluator. Each answer pair undergoes two evaluation rounds with position inversion to mitigate position bias. Only judgments consistent across both rounds (A-A, B-B, or T-T) are counted as valid. We enforced 500-token response limits to mitigate verbosity bias and did not address self-enhancement bias (both responses use the same base model).

The judge prompt emphasizes: relevance, accuracy, depth, and creativity; natural conversational tone; harsh penalties for fabricated information; and objectivity independent of response order or length.

**RAGAs Framework (Es et al., 2023)**: This reference-free evaluation assesses three dimensions:

1. **Faithfulness**: Do generated responses adhere to retrieved context? Score = proportion of supported claims.
2. **Context Relevance**: Does retrieval surface pertinent information? Score = relevant / total sentences.
3. **Answer Relevance**: Is the generated response appropriate to the query? Measured via cosine similarity between generated questions and original query.

These metrics complementarily assess retrieval quality (Context Relevance), grounding (Faithfulness), and generation quality (Answer Relevance).

### 5.3 Efficiency Results (RQ1)

Our system achieved an average prompt generation time of 0.52 seconds—a critical threshold for natural conversation. This timing enables real-time educational deployment while achieving comparable latency to generation itself (GPT-4: ~5-7 tokens/second). Multi-stage reflection systems requiring 3-5 sequential LLM calls would incur 15-30+ seconds of latency, fundamentally precluding conversational interaction. The average prompt generation time of 0.52 seconds across hundreds of test interactions, measured from query receipt to final prompt delivery to the LLM. This timing includes: query embedding (1 API call), parallel similarity search across both RAG indexes, two-stage retrieval (description to full memory +

metadata), prompt template population and formatting, and query-response pair embedding for conversational memory (1 API call).

**Contextualization**: To assess whether this speed enables real-time interaction, we compare it against theoretical LLM generation speeds. GPT-4 generates approximately 5-7 tokens/second; GPT-3.5 generates 12-15 tokens/second. In 0.52 seconds, GPT-4 would generate approximately 3-4 words while GPT-3.5 would generate approximately 6-8 words.

Our system constructs a 1,500-token context (3 rich memories + metadata + conversation history) in the time it would take GPT-4 to generate a single short phrase. This demonstrates that retrieval and prompt assembly contribute negligible latency compared to generation itself, the hallmark of real-time viability.

**Comparison to Multi-Stage Systems**: Self-reflection architectures requiring 3-5 sequential LLM calls (initial generation, reflection, retrieval, regeneration) would require 15-30+ seconds for comparable complexity, assuming approximately 3-5 seconds per LLM call. Iterative RAG chains with recursive refinement similarly accumulate latency linearly with depth. Our parallel, pre-processed approach eliminates this sequential dependency.

**Energy Implications**: Beyond latency, our architecture reduces energy consumption by minimizing LLM API calls. Traditional systems might invoke the LLM 5-10 times per query (multiple retrievals, reflections, refinements); we invoke it twice: once for the main response, once for embedding the query-response pair. This 5-10x reduction in API calls translates directly to proportional energy savings, particularly relevant for high-traffic educational deployments.

## 5.4 Comparative Quality Evaluation (RQ2)

### 5.4.1 LLM-as-a-Judge Results

Table 2 presents pairwise comparison results across our ablation studies and baseline systems.

| Comparison | Our System Wins | Baseline Wins | Ties | Judgments |
|---|---|---|---|---|
| **Ablation Studies** | | | | |
| Full vs. Static Only | 23 | 4 | 6 | 33 |
| Full vs. Static + Episodic | 13 | 3 | 17 | 33 |
| **Cross-Model RAG Comparison** | | | | |
| Full vs. Trad RAG (GPT-4) | 6 | 6 | 21 | 33 |
| Full vs. Trad RAG (GPT-3.5) | 15 | 8 | 10 | 33 |
| Full vs. Trad RAG (GPT-3) | 15 | 6 | 12 | 33 |

**Table 2**: LLM-as-a-judge pairwise comparison results. "Full" denotes complete system with episodic memory and affective-semantic metadata.

**Ablation Study Interpretation**: The full system substantially outperforms static-only prompting (23-4-6), confirming that episodic memory retrieval provides substantial value. The comparison against Static + Episodic (13-3-17) shows more modest improvement with many ties, suggesting that affective-semantic metadata provides incremental benefit but episodic structure itself delivers primary impact.

**Model Scale Effects**: Results reveal a critical pattern: our structured episodic approach particularly benefits smaller models. With GPT-4, our system achieved parity with traditional RAG (6-6-21), with most responses judged equivalent. We interpret this as GPT-4's superior internal knowledge compensating for traditional RAG's less structured context. The model's extensive training on Van Gogh materials enables high-quality responses even from fragmented biographical chunks. However, with GPT-3.5, our system demonstrated clear superiority (15-8-10), nearly doubling win rate over traditional RAG. This advantage amplified further with GPT-3/Davinci-002 (15-6-12), showing that structured, first-person episodic memories with explicit metadata provide greater benefit as base model capability decreases.

**Implications**: These results directly support our target deployment scenario. While large-scale models achieve acceptable performance with various RAG approaches, smaller models suitable for edge deployment, educational institutions, and resource-constrained environments show substantial improvement with our structured memory architecture. This pattern suggests our methodology's primary value lies in democratizing character-based dialogue systems by enabling good performance from accessible models rather than requiring expensive, high-capability cloud APIs.

### 5.4.2 RAGAs Framework Results (RQ3)

Table 3 presents RAGAs metrics across model scales, comparing our system against traditional RAG using identical static contexts. Higher values indicate better performance (scale: 0–1).

| Model | System | Faithfulness | Answer Relevance | Context Relevance |
|---|---|---|---|---|
| **GPT-4** | Our System | 0.73 | 0.88 | 0.16 |
| | Traditional RAG | 0.58 | 0.88 | 0.13 |
| **GPT-3.5** | Our System | 0.76 | 0.88 | 0.17 |
| | Traditional RAG | 0.66 | 0.88 | 0.11 |
| **GPT-3** | Our System | 0.36 | 0.85 | 0.18 |
| | Traditional RAG | 0.19 | 0.81 | 0.13 |

**Table 3.** RAGAs evaluation scores across model scales. Higher values indicate better performance (scale: 0-1)..

Our system consistently demonstrates superior faithfulness (the degree to which generated responses adhere to retrieved context rather than hallucinating or diverging). The advantage ranges from +0.15 for GPT-4 to +0.17 for GPT-3, showing that structured first-person memories with explicit temporal-social-emotional framing help models ground their responses regardless of base capability. Our retrieval mechanism consistently surfaces more pertinent information (+0.03 to +0.07 advantage). The two-stage retrieval approach (using concise augmented contexts for similarity matching but providing rich narratives for generation) appears to improve precision. Traditional RAG retrieving directly from raw biographical chunks often includes tangentially related material, diluting relevance. Both systems achieve comparable Answer Relevance scores (0.81-0.88), with minimal differences. This metric primarily reflects LLM generation capability rather than retrieval architecture, explaining the parity. The slight advantage our system shows on GPT-3 (0.85 vs 0.81) may indicate that clearer, more structured context helps weaker models stay on-topic. The apparent contradiction (GPT-4 parity in judge evaluations but RAGAs superiority) resolves when we recognize what each measures. RAGAs assesses retrieval and grounding quality in isolation; LLM-as-a-judge evaluates end-to-end response quality including fluency, completeness, and conversational naturalness. GPT-4's strong internal knowledge allows it to produce high-quality responses even with less faithful grounding by drawing on training data. Smaller models lacking this knowledge reservoir depend more heavily on retrieved context quality, explaining why structured episodic memory provides greater benefit at smaller scales.

## 5.5 Intermediate Conversational Memory and Hallucination Resistance

We qualitatively assessed our intermediate memory mechanism through extended conversations examining thematic continuity. Observational testing confirmed the intended behavior: after discussing Van Gogh's relationship with nature and art, a query about "favorite painting brush" naturally referenced the earlier discussion and retrieved related biographical experiences about Theo's role in artistic development. This demonstrates the system retrieving conversational memory, using it to query long-term episodic memory, and weaving results together naturally. Traditional RAG systems lacking conversational memory would treat each query independently, losing this thematic coherence.

While we did not conduct formal hallucination quantification, we performed regex pattern searches for third-person perspective drift, a hallucination type observed in preliminary experiments with traditional RAG. Specifically, we searched for patterns like "Van Gogh [verb]", "he [verb]", and similar third-person constructions in generated responses.

**Results**: Zero instances found in our system's outputs across approximately 200 test conversations. In contrast, preliminary traditional RAG tests showed occasional third-person drift despite explicit prompt instructions against it. We attribute this to our first-person memory structure. Providing LLMs with consistent first-person exemplars appears to reinforce perspective maintenance more effectively than prompt instructions alone.

The system also successfully handled the "uneventful date" query (asking about a random date with no significant events) by acknowledging the absence of specific memories rather than fabricating events, though we did not systematically test this across multiple such queries.

Our evaluation provides evidence for efficiency gains and quality improvements but has several limitations. All evaluations employed automated metrics (LLM-as-a-judge, RAGAs) or computational measurements (timing) rather than human user studies. We evaluated only the Van Gogh implementation, and while the architecture generalizes by design, empirical validation with other historical figures would confirm methodology transferability. Our hallucination assessment focused on perspective consistency rather than comprehensive factual verification. While RAGAs Faithfulness measures grounding in retrieved context, it doesn't verify that the retrieved context itself is factually accurate. Our manual review corrected 14 date hallucinations, suggesting less than 1% error rate, but comprehensive verification would require extensive expert annotation. These limitations suggest clear directions for future work while not undermining the core technical contributions: demonstrable efficiency gains, consistent quality improvements on smaller models, and a generalizable architecture for character-based dialogue systems.

# 6    APPLICATIONS AND VISUALIZATION

Our architecture serves two co-equal purposes: enabling natural conversational interaction with historical character agents and providing analytical tools for exploring the cognitive-experiential landscape of historical figures. These applications share the same underlying structured episodic memory substrate but address different use cases.

**Conversational Dialogue Application:** The primary deployment scenario involves educational settings where students, museum visitors, or general audiences engage with historical figures through natural language dialogue. Our system's efficiency (0.52s prompt generation) and token constraints (2k total) make it suitable for deployment on classroom computers, museum kiosks, or modest cloud infrastructure. Here is an example biographical query:

*User*: "Did you ever fight with your brother?"

*Van Gogh*: "Yes, there were times when Theo and I clashed deeply. In December 1881, Theo sent me a scolding letter calling me a coward. His words cut deep. Despite our tensions around finances and my relationship with Sien, he remained my closest confidant. Our bond was complex, both deeply connected and occasionally strained."

The intermediate conversational memory mechanism enables thematic coherence across multi-turn dialogues. When asked about painting brushes after earlier discussing nature and art, the system naturally referenced the prior conversation: "Reflecting on my experiences, as I mentioned earlier in our conversation, my deep connection with nature, art, and the human soul has always guided my hand." This demonstrates retrieval of conversational memory, using it to query long-term episodic memory, and weaving results together naturally. The system's grounding in structured biographical data also enables appropriate responses to queries about undocumented periods, acknowledging the absence of specific memories while providing relevant temporal context rather than fabricating events.

**Memory Exploration and Visualization:** Our affective-semantic metadata enrichment transforms episodic memories into a multidimensional analytical substrate supporting novel approaches to biographical investigation.

**Spatiotemporal Heatmaps**: By plotting memories on world maps using geocoded coordinates, we visualize the geographic scope of a figure's experiences. For Van Gogh, heatmaps reveal dense clustering in the Netherlands (childhood), significant concentration in Paris (artistic development), extended periods in southern France (Arles, Saint-Rémy), and sparse coverage in Belgium (Borinage period). Interactive filtering by date range or emotional valence enables queries like "Where did Van Gogh's most positive experiences occur?" offering immediate insights into mobility patterns and the relationship between location and life phases.

**Emotional Trajectory Analysis**: Plotting average emotional valence and arousal across Van Gogh's lifespan reveals biographical patterns that might remain hidden in narrative accounts (Figure 4). Pronounced negative valence spikes in 1880, 1885, and 1890 correspond to known biographical crises: beginning his uncertain artistic career, creating "The Potato Eaters" amid personal struggles, and his final year marked by mental health deterioration. Arousal decline in final years suggests emotional exhaustion alongside creative intensity. The 1874 positive spike corresponds to family Christmas memories, a rare oasis of contentment. These patterns emerge organically from aggregating memory-level emotional annotations, demonstrating how quantitative analysis of qualitative biographical materials can surface temporal structures worthy of deeper investigation.

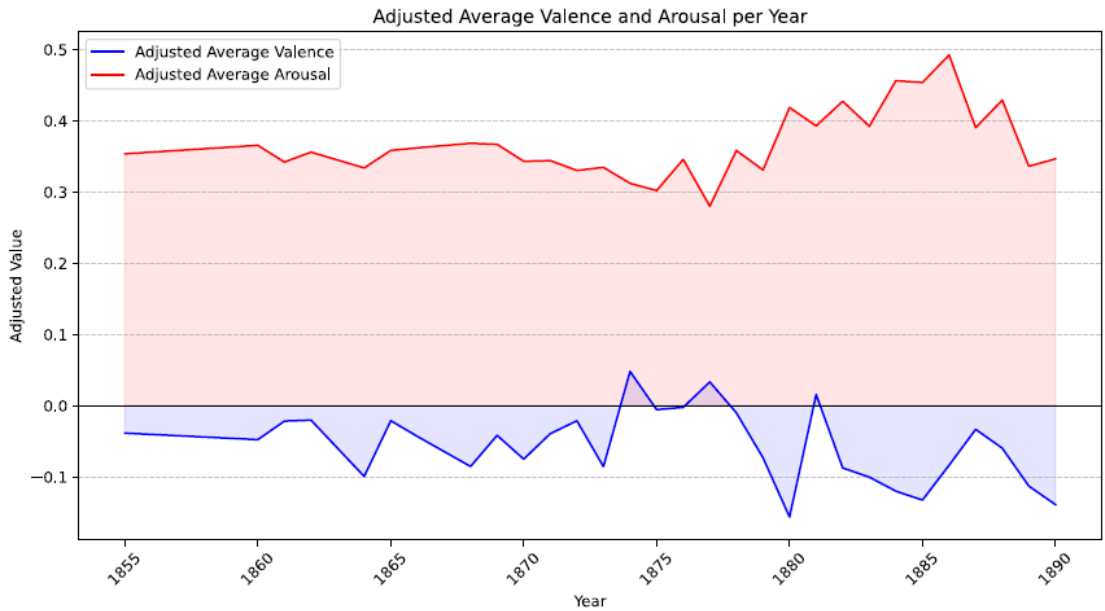

**Figure 4**. Yearly Weighted Average Valence and Arousal over Van Gogh's life.

**Social Network Analysis**: Tallying character mentions reveals relationship centrality and salience. In Van Gogh's dataset, Theo van Gogh appears in more than 400 memories (22% of dataset), reflecting his central role as brother, confidant, and financial supporter. Paul Gauguin shows significant presence (80+ memories) despite brief actual cohabitation, indicating emotional and artistic impact. Romantic interests (Loyer, Vos, Hoornik, Begemann) appear as distinct temporal clusters, each representing intense but temporally bounded emotional focus periods. This visualization approach offers researchers tools for investigating questions like "When did Van Gogh's relationship with Gauguin transition from artistic admiration to personal conflict?"

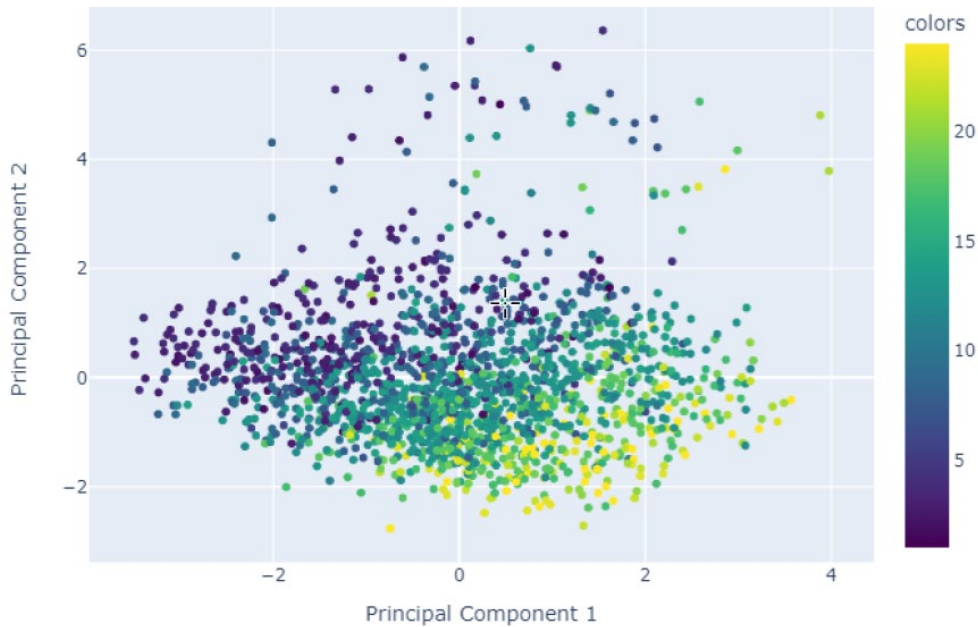

**Figure 5.** PCA of the Autonoetic Memory. Yellow memories: higher valence, dark-blue: lower valence.

We applied PCA to the combined feature space (embeddings plus numerical valence, arousal, relevance) to project the 1,774 memories into 2D space for visualization (Figure 5). Color-coding by valence reveals clustering: higher valence memories (yellow) concentrate in one region, while lower valence memories (dark blue) cluster elsewhere, suggesting that certain memory configurations (particular combinations of characters, locations, topics) correlate with emotional tone. Researchers can isolate memory subsets for focused analysis by filtering for specific characters, time periods, or emotional ranges.

As users engage in dialogue, we record which memories the system retrieved for each query. Plotting these retrieval sequences in the PCA-reduced space reveals the trajectory of conversation through Van Gogh's experiential landscape (Figure 6). The path visualization provides: for educators, awareness of what experiential territory a conversation has covered; for students, metacognitive reflection on their question-asking patterns; for researchers, insight into how different query strategies surface different memorial configurations. This treats biographical exploration as navigation through cognitive-experiential space rather than information retrieval from a database.

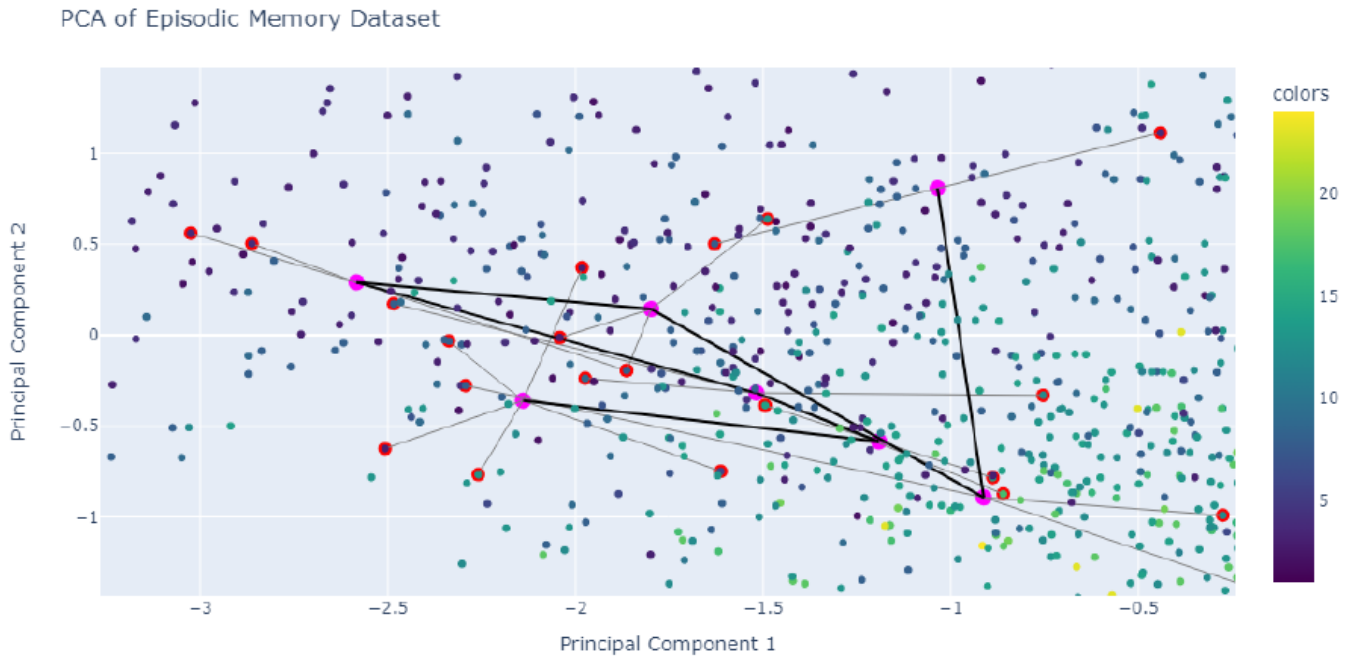

**Figure 6.** Conversation Path through memory space during one interaction.

The power of our approach emerges from treating dialogue and visualization as complementary rather than separate. A student's conversation generates a path through memory space visible in the visualization tool. An educator exploring emotional trajectory patterns might notice interesting memories, then switch to dialogue mode for elaboration. The visualization tools serve practical needs: educators can identify memory clusters representing key biographical periods for lesson planning, researchers could compare experiential patterns across multiple historical figures, and museums could deploy dual interfaces with conversational kiosks for visitor engagement and visualization terminals for deeper exploration. By implementing both capacities in our architecture, we create systems that provide windows into the structure of historical consciousness as documented in biographical materials.

# 7  DISCUSSION

We demonstrate that principled architectural design resolves the efficiency-accuracy dilemma in character-based dialogue systems can be resolved through principled separation of computational concerns: conducting expensive operations offline during dataset construction while optimizing runtime processes for speed. This section synthesizes our findings, examines implications, and acknowledges limitations.

## 7.1 Addressing Core Research Goals

Our architecture successfully addresses the three research goals established in the introduction.
**Goal 1 (Robust Episodic Memory Datasets)**: Our multi-stage augmentation pipeline transforms unstructured biographical materials into structured, queryable episodic memories enriched with affective-semantic metadata that enable both rapid retrieval for dialogue and analytical visualization.
**Goal 2 (Efficient Real-Time Interactions)**: The 0.52-second average prompt generation time establishes real-time viability, enabling deployment on classroom computers, museum kiosks, and modest hardware while achieving proportional energy savings through reduced API calls.
**Goal 3 (Performance Benchmarking)**: Our evaluation reveals that while achieving parity with traditional RAG on GPT-4, the true advantage emerges with smaller models (GPT-3.5 and GPT-3), suggesting that episodic memory architecture democratizes character embodiment by making smaller, more accessible models viable for quality performance.

## 7.2 Implications for Sustainable and Accessible AI

Beyond technical contributions, our work engages with pressing concerns about AI's environmental impact and accessibility barriers. The reduction from 5-10 LLM calls per query to 2 calls translates directly to proportional energy savings. For a system serving 10,000 daily queries (modest for educational applications), this represents thousands of avoided LLM invocations. Each consumes energy for GPU computation, data center cooling, and network transmission. By demonstrating that smaller models perform well with appropriate architectural support, we challenge the implicit assumption that quality character embodiment requires expensive cloud APIs. This has equity implications: well-resourced institutions can afford unlimited GPT-4 API calls, but underfunded schools, small museums, and researchers in developing regions need alternatives. Our approach suggests that thoughtful system design can partly compensate for raw model capability, expanding access to these technologies. Conversations with educators during development revealed practical constraints often overlooked in research: limited budgets, restricted internet access, privacy concerns about cloud APIs, and aging hardware. With these issues in  mind, while we demonstrated benefits for smaller proprietary models (GPT-3.5, GPT-3), validating performance with fully open models deployable on consumer hardware further confirms our democratization claims. Our final post-study validation deployed our system using a small local open source model,  Qwen 2.5:14B,  on a consumer-grade RTX 4080 (12 GB VRA) laptop. The system maintained functional accuracy with practical performance metrics—~3 s average latency (vs 1–2 s cloud GPT-4) with these additional advantages:

  - No recurring API or subscription costs after initial setup
  - Unlimited offline usage for classrooms, museums, and remote installations
  - Full data privacy and security, since all processing remains on-device with no cloud transmission
  - Reduced environmental footprint: smaller, efficiently models avoids energy-intensive cloud inference

A mid-range PC can host the system indefinitely, transforming cloud-based operational expenses into a one-time investment. Responses remained conversationally natural and factually grounded. This profile—sub-5 s responses, zero ongoing cost, offline capability, privacy, and sustainability—validates the architecture's accessibility goal: structured episodic memory enables character embodiment without dependence on expensive, energy-intensive cloud APIs.

### 7.3 Limitations and Boundary Conditions

Our work has several important limitations. Despite architectural generalizability, we evaluated only Van Gogh. This figure offers advantages (extensive documentation, LLM training coverage) that may not apply to others. Our pipeline assumes biographical sources of Van Gogh are accurate. However, Naifeh-Smith's biography, while critically acclaimed, represents interpretation of historical evidence, not unmediated fact. Van Gogh's letters exist in English translation; the biography is English-language. Multilingual figures or those whose materials resist translation pose challenges our methodology doesn't address. Beyond Van Gogh, our historical character procedure needs substantial textual materials, a minimum of approximately 500 pages of biographical content. This excludes many historically important but sparsely documented individuals. Historical figures from distant eras (ancient world, medieval periods) present challenges. Our approach likely works best for 18th-20th century Western figures, a significant constraint.

All quality evaluation employed automated metrics. Direct user testing with historians, educators, and students would validate practical utility. We demonstrated resistance to third-person perspective drift but didn't systematically measure factual accuracy. While RAGAs faithfulness suggests improved grounding, comprehensive fact-checking against biographical ground truth would provide stronger evidence.

### 7.4 Broader Implications for HCI and Conversational AI

Our work contributes to several ongoing conversations in human-computer interaction and artificial intelligence. Traditional RAG systems treat knowledge as a database to query for facts. Our episodic memory approach treats it as experiential territory to explore, a fundamentally different interaction metaphor. This shift from "ask questions, get answers" to "navigate cognitive-experiential space" suggests new design patterns for AI systems meant to convey understanding rather than merely retrieve information. The transparency enabled by structured metadata (knowing why a memory was retrieved, seeing its emotional context, understanding its autobiographical significance) offers a path toward explainable AI that doesn't rely on post-hoc rationalization. Treating real-time responsiveness as a hard requirement drove architectural decisions that yielded broader benefits (energy efficiency, accessibility, edge deployment viability). This suggests that constraint-driven design (deliberately limiting resources to force creative solutions) can produce systems with emergent advantages beyond the targeted constraint. Our work demonstrates productive engagement between cognitive science theory and engineering practice: neither blind adherence to biological fidelity nor complete disregard for memory research, but selective adoption of concepts that provide actionable architectural guidance.

## 8 CONCLUSION AND FUTURE WORK

Character-based dialogue systems face a fundamental tension: achieving conversational depth and factual accuracy typically requires computational approaches incompatible with real-time interaction. Multi-stage reflection systems deliver rich, grounded responses but incur latency that disrupts natural conversation, while simple retrieval methods sacrifice the experiential richness essential to authentic character embodiment. This work demonstrates that principled architectural design can resolve this dilemma. By conducting computationally expensive operations once during offline dataset construction (transforming biographical materials into structured episodic memories enriched with affective-semantic metadata) and optimizing runtime operations for speed through parallel retrieval, we achieve sub-second prompt generation (0.52s average) while maintaining response quality exceeding traditional RAG approaches.

Our evaluation reveals a critical finding: structured episodic memory particularly benefits smaller language models, showing substantial quality improvements with GPT-3.5 and GPT-3 level models while achieving parity with traditional RAG on GPT-4. This pattern suggests our architecture's primary value lies in democratizing character embodiment by enabling good performance from accessible, deployable models. For educational institutions, museums, and research environments with computational constraints, this accessibility proves essential. The architecture supports dual applications of equal importance. Conversational dialogue enables natural language interaction for educational engagement, while memory visualization tools transform biographical data into navigable cognitive-experiential landscapes supporting curriculum development, biographical research, and historical analysis. This dual-use design (treating the structured episodic memory dataset as both a retrieval substrate for generation and an analytical space for exploration) distinguishes our approach from purely conversational systems.

The methodology generalizes beyond Vincent van Gogh to any historical figure with substantial textual documentation combining biographical narratives and first-person materials. Our multi-stage augmentation pipeline (perspective transformation via screenplay generation, systematic information extraction, affective-semantic enrichment, and quality assurance) requires only substituting source materials and character-specific prompts. No domain knowledge or figure-specific fine-tuning is necessary.

**Future work:**

Three  research directions would strengthen and extend this work. First, formal user evaluation with educators, students, historians, and museum visitors would validate educational efficacy and reveal authentic usage patterns. Systematic fact-checking against biographical sources would quantify hallucination rates beyond the perspective consistency we assessed.

Second, empirical generalizability testing requires implementing systems for diverse historical figures varying in documentation density, cultural background, historical period, and domain expertise. This would reveal which architectural elements require domain-specific adaptation and establish boundary conditions for the methodology.

Finally, multimodal integration would extend the architecture beyond text. Linking memories to artworks, photographs, or historical documents would enable visual queries ("show me paintings from your happiest period") and cross-modal analysis correlating artistic output with emotional trajectories. This naturally extends to other figure types: scientists' equations, political figures' speeches, or authors' manuscripts, each offering domain-specific modalities beyond biographical text.

As large language models continue evolving, the tension between capability and accessibility will intensify. Our work suggests that architectural innovation (offline augmentation, structured memory, parallel retrieval, cognitive scaffolding through metadata) offers a complementary path forward. Rather than relying solely on ever-larger models, we can design systems that amplify smaller models' capabilities through principled information organization and efficient access patterns.

For researchers, educators, and developers seeking to create character-based dialogue systems that are simultaneously accurate, responsive, accessible, and analytically rich, our architecture offers both a working solution and a design philosophy: separate computational concerns temporally, structure information cognitively, optimize for the constraints that matter most, and recognize that conversational interaction and contemplative exploration serve complementary purposes in understanding historical consciousness.